\documentclass{llncs}

\usepackage{url}
\usepackage{graphicx}

\usepackage{floatrow}
\begin{document}

\mainmatter

\title{Integrating Know-How \\into the Linked Data Cloud}

\author{Paolo Pareti\inst{1}\inst{2} \and Benoit Testu\inst{1} \and
Ryutaro Ichise\inst{1} \and \\ Ewan Klein \inst{2} \and Adam Barker \inst{3}}

\institute{National Institute of Informatics, Tokyo, Japan,\\
\email{p.pareti@sms.ed.ac.uk, benoit.testu@u-psud.fr, ichise@nii.ac.jp}
\and
University of Edinburgh, Edinburgh, United Kingdom\\
\email{ewan@inf.ed.ac.uk}
\and
University of St Andrews, St Andrews, United Kingdom\\
\email{adam.barker@st-andrews.ac.uk}
}

\maketitle

\begin{abstract}
This paper presents the first framework for integrating procedural knowledge, or ``know-how'', into the Linked Data Cloud. 
Know-how available on the Web, such as step-by-step instructions, is largely unstructured and isolated from other sources of online knowledge. To overcome these limitations, we propose extending to procedural knowledge the benefits that Linked Data has already brought to representing, retrieving and reusing declarative knowledge. 
We describe a framework for representing generic know-how as Linked Data and for automatically acquiring this representation from existing resources on the Web.
This system also allows the automatic generation of links between different know-how resources, and between those resources and other online knowledge bases, such as DBpedia. We discuss the results of applying this framework to a real-world scenario and we show how it outperforms existing manual community-driven integration efforts. 


\end{abstract}

\section{Introduction}

\par The Web contains a large amount of procedural knowledge, or \emph{know-how}, in many domains of human interests, ranging from cooking recipes to software tutorials and social skills. As such, it has become one of the major sources of knowledge for anybody who is interested in performing a task. Online knowledge also has the potential for helping machines to understand and reason over common-sense human activities \cite{Myaeng2013ExperientialKnowledgeMining}. However, the potential for applying this knowledge is severely restricted due to its lack of structure, the diversity of representation formats and its isolation from other knowledge sources. In this context, we argue that Linked Data is an ideal representation for overcoming these restrictions on using know-how at web-scale. The main contributions of this paper are:
\begin{itemize}
	\item The description of the first framework that can automate the creation and the integration of know-how into the Linked Data Cloud.
	\item The validation of this framework with a concrete implementation which outperforms existing know-how integration efforts.
\end{itemize}
The main benefit of a formal representation of human know-how is to allow machines to better understand human processes, making them reusable in different applications. We adopt a generic definition of the term process which includes any entity which has the potential for being performed, such as step-by-step instructions. 

\par We present a framework that overcomes the limitations of the existing know-how and enables it to be integrated into the Linked Data Cloud. This framework can be divided in two components:
\begin{itemize}
	\item Knowledge Extraction and Representation. This framework allows the automatic formalization of existing know-how based on a generic and lightweight Linked Data vocabulary. This component addresses the limitation of the lack of explicit structure and shared semantics.
	\item Linked Data Integration. This framework can automatically discover links between processes and other existing Linked Data. This component addresses the problem of the isolation of individual processes from related knowledge.
\end{itemize}
This framework has been implemented and validated in a real-world scenario. We have applied it to two different large-scale know-how repositories and created a Linked Data representation of over 200,000 processes. These processes have been integrated both with each other and with DBpedia \cite{Auer2007DBpedia}. We have compared the quality of the results with existing user-generated links and we have tested them in a practical application.

\section{Problem Formulation} \label{problemFormulation}
The problem addressed by this paper is the \emph{effective} integration of a particular kind of procedural knowledge, that we call \emph{human know-how}, into the Linked Data Cloud. The integration will be considered effective if the quality of the generated links can be shown superior to a real-world benchmark. We define human know-how as the procedural knowledge that involves humans as the main actors. Other well-known types of procedural knowledge, such as programming languages or business workflows, are excluded by this definition as they are meant to be executed by artificial agents such as computers. The formal representation of human know-how presents some unique challenges:
\begin{itemize}
	\item Knowledge can be vague, erroneous or missing. For example, the details on how to perform a step might not be specified.
	\item Knowledge is distributed across different repositories on the Web.
	\item Knowledge is in constant evolution. New processes can be defined and existing processes can be modified.
\end{itemize} 
Considering these challenges, we propose two requirements that are necessary to make human know-how machine understandable.
The first requirement is that the knowledge representation language needs to be generic and lightweight. A generic representation is required because human know-how covers many different domains. This representation also needs to be lightweight in order to avoid inconsistencies or wrong inferences when integrating conflicting or erroneous information from distributed sources. This does not exclude the possibility of adopting other logic-heavy representations if required for specific applications.
\par The second requirement is that it should be possible to automatically generate structured knowledge about a process from the unstructured user-generated representation. A manual approach is impractical for two reasons: first, because of the large number of existing know-how resources and second, because of their evolution over time, which would require constant revisions. 

\section{Related Work}

\subsection{Procedural Knowledge Representation} \label{relatedWorkKnowledgeRepresentation}
There is a rich body of research into methods of representing and reasoning with procedural knowledge, for example in the Automated Planning and Problem Solving Methods fields. These systems, however, are not sufficiently lightweight and generic to conveniently represent human know-how. Logic-heavy representations, such as OWL-S \cite{Martin2004owls} and the Process Specification Language (PSL) \cite{Gruninger2003PSL}, require the information about a process to be complete and correct. Their application in the human know-how domain is therefore inconvenient, as it would constantly face the problem of inconsistencies and wrong inferences.
\par Another limitation of existing languages is domain specificity. OWL-S, for example, defines a process as a ``specification of the ways a client may interact with a service" \cite{Martin2004owls}. This definition is not compatible with a more general interpretation of a process which might not involve neither clients nor services. Lastly, languages might not be sufficiently expressive. For example, the vocabulary defined by Schema.org\footnote{\url{http://schema.org/}} currently lacks relations to define the decomposition of a process into steps. The arguments that we have made to show the limitations of OWL-S, PSL and Schema.org can also be applied to other existing languages.

\subsection{Human Know-How Extraction} \label{relatedWorkKnowledgeExtraction}
Several research projects have already attempted to extract human know-how from manually generated instructions. One of the most frequent approaches, based on Natural Language Processing (NLP), has been used to extract knowledge from domain-independent know-how repositories, such as wikiHow\footnote{\url{http://www.wikihow.com/}} \cite{Addis2011FromUnstructured}, \cite{Jung2010AutomaticConstruction} as well as domain-specific ones, like the medical domain \cite{Song2011ProceduralKnowledge}. An approach based on statistical analysis has also been used to extract procedural knowledge from a more diverse set of Web documents not necessarily focused on know-how \cite{Fukazawa2010AutomaticModeling}.
\par These approaches were used to obtain a deep logical understanding of the processes which resulted in a loss of accuracy, as most user-generated instructions are inherently vague and cannot be analyzed reliably. Our system, instead, does not require a particular level of detail in the formalization. For this reason, the structure of a process is extracted only when this can be done with confidence. This situation occurs when such structure already exists, for example when the steps of a process have been clearly divided into an ordered list.

\subsection{Applications of Human Know-How}
A machine understandable representation of human know-how can be applied in a wide variety of areas ranging from Information Retrieval to Service Recommendation \cite{Myaeng2013ExperientialKnowledgeMining}. Two notable applications will be discussed here, namely Activity Recognition and process automation. The goal of Activity Recognition is to identify a top level activity (or intention) from a set of observations \cite{Eunju2010HumanActivity}. The intention of preparing tea, for example, could be inferred after observing an agent boiling water and placing a tea bag in a cup. A challenge faced by Activity Recognition systems is the acquisition of a model of the activities to be recognized. This model can be extracted from human know-how on the Web and used to recognize common human activities \cite{Perkowitz2004MiningModels}.
\par Another notable application of formalized human know-how involves the automation of activities. One experiment that attempted this kind of automation employed a robotic agent \cite{Tenorth2011WebEnabled}. This agent attempted to perform the activity of preparing a pancake by following user-generated instructions retrieved from a wikiHow website. This experiment explored the potential of human know-how for process automation and highlighted the importance of integrating individual processes with external sources of knowledge. Artificial agents, in fact, require more information about a process than what can typically be extracted from a single set of instructions, as they lack human common sense. Knowledge about an ingredient, for example, allowed the agent to learn what it looked like and whether, by virtue of being perishable, it might be found in the refrigerator.

\begin{figure}[tb]
	\caption{Diagram of the human know-how integration framework}
			\centering
			\includegraphics[width=1.00\textwidth]{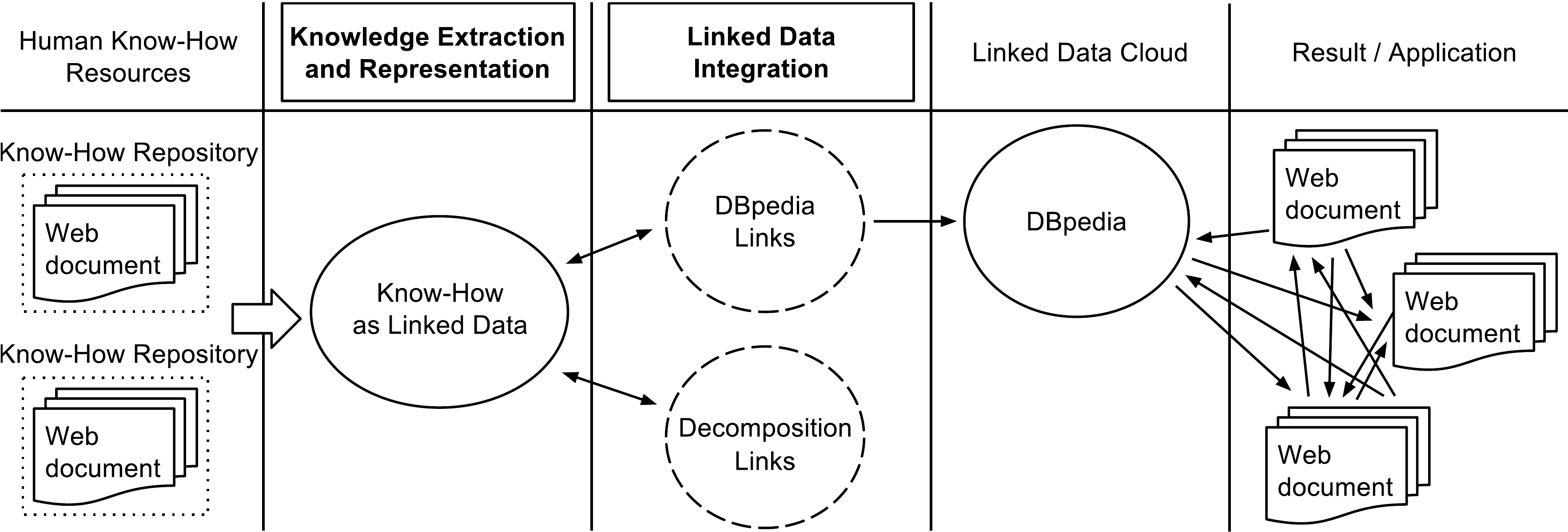}
	\label{fig:integrationDiagram}
\end{figure}

\section{Methodology}
The integration of human know-how into the Linked Data Cloud requires solving a number of different problems. The most important of those are the representation of know-how as Linked Data and the generation of links to external sources of information. Our framework addresses each of those issues with a different component. Figure \ref{fig:integrationDiagram} schematizes the general workflow of our system. The input of the system is a set of human know-how resources. These resources are analyzed by the first component of our framework, namely \textbf{Knowledge Extraction and Representation}, and converted into Linked Data. This Linked Data representation is integrated with the Linked Data Cloud by the second component of our framework: \textbf{Linked Data Integration}. This last component generates two kinds of links, namely links to DBpedia (DBpedia links) and links between processes (decomposition links). More details about these two components will be provided in the next subsections.

\subsection{Knowledge Extraction and Representation} 
In section \ref{relatedWorkKnowledgeRepresentation} we discussed several issues in reusing existing knowledge representation languages in the human know-how domain. This lead us to the development of a Linked Data vocabulary which is both lightweight and generic \cite{Pareti2014}.  This vocabulary is sufficient to represent the two main concepts that can be reliably extracted from semi-structured human know-how. These concepts, namely dependencies and process decompositions, play a key role in most procedural knowledge representation formalisms. This vocabulary is based on just three properties, as shown in Table \ref{tab:vocabulary}.
\begin{table}[tp]
	\centering
		\begin{tabular}{ | l | l |}
			\hline
			Prefix & Namespace \\ \hline
			prohow: & \url{http://vocab.inf.ed.ac.uk/prohow#} \\ \hline \hline
			Term & Definition when $X$ is the subject and $Y$ is the object \\ \hline
			\url{prohow:has_step} & $Y$ can help accomplishing/obtaining $X$  \\  \hline
			\url{prohow:has_method} & $Y$ can be accomplished/obtained instead of $X$ \\ \hline
			\url{prohow:requires} & $Y$ should be accomplished/obtained before doing $X$ \\ \hline
		\end{tabular}
	\caption{The vocabulary to represent processes}
	\label{tab:vocabulary}
\end{table}

\begin{description}
\item[\texttt{prohow:has\_step}] This property can be used to decompose a complex process into its various sub-processes. For example, this property could connect the process ``make a pancake'' with its step ``mix the ingredients''.  
\item[\texttt{prohow:has\_method}] This property can connect a process with an alternative way of achieving it. For example, this property could connect the process ``make a pancake'' with the more specific process ``make a lemon pancake''. 
\item[\texttt{prohow:requires}] This property defines a dependency between two entities. The process ``put the mix on a pan'', for example, depends on the process ``mix the ingredients'', which should be done in advance. 
\end{description}
These relations can also be used to connect processes with objects. In our example, the process ``put the mix on a pan'' could specify the object ``pancake mix'' as a requirement using the \url{prohow:requires} relation. In the human know-how domain, the distinction between processes and objects is often vague.

\subsubsection{Knowledge Extraction from Know-How Repositories.} \label{subsec:comp1extraction}
\begin{figure}[tb]
	\caption{Extraction of the Linked Data representation of human know-how}
			\centering
			\includegraphics[width=1.00\textwidth]{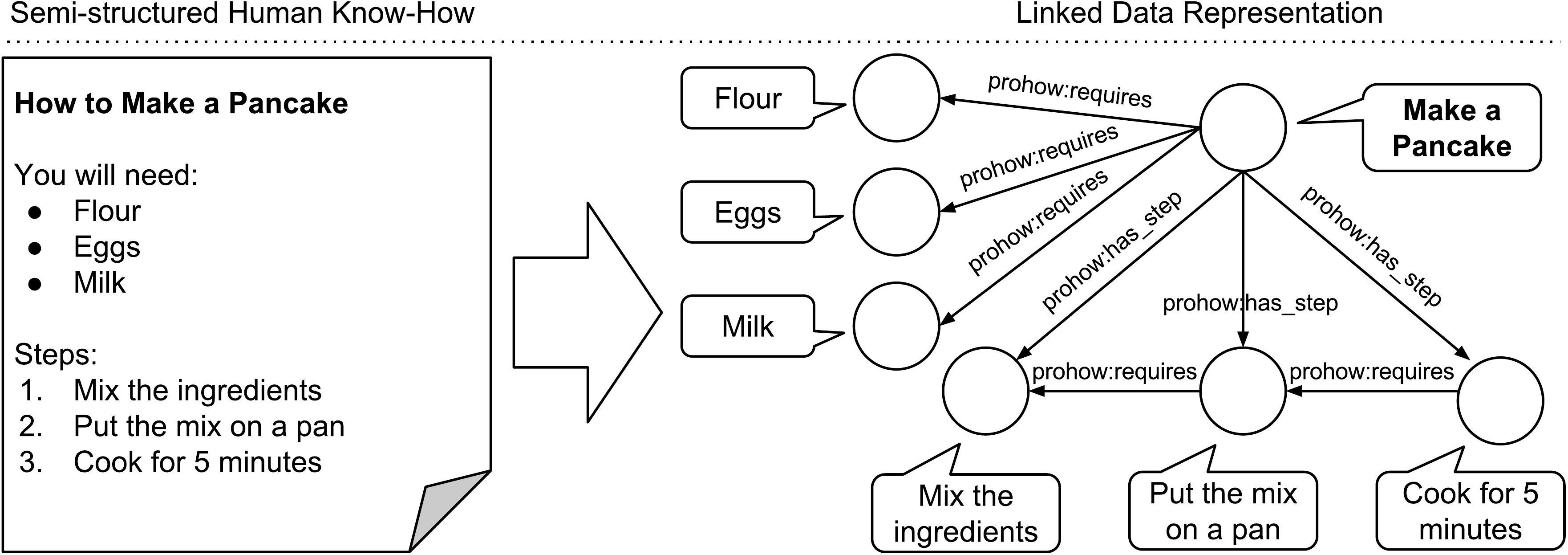}
	\label{fig:ExtractionDiagram}
\end{figure}
Our approach to knowledge extraction is focused on semi-structured resources. Examples of those resources can be found in the wikiHow and Snapguide\footnote{\url{http://snapguide.com/}} websites. The structure of those resources typically contains: (1) a title denoting the main task that the process achieves, (2) the category of the process, (3) a list of the distinct requirements of the process, (4) a hierarchical structure of the steps of the process and (5) the order in which the various steps should be performed. 
Our approach extracts this structure to reliably decompose a process into a number of entities and relations. A simplified example of such extraction is depicted in Figure \ref{fig:ExtractionDiagram}. Each extracted entity is given a unique URI and it is linked with the other entities of the same process. Finally, each entity is connected with its human-understandable representation using the Open Annotation Data Model.\footnote{\url{http://www.openannotation.org/}} 
\par The main advantages of our approach are two. First, our extraction is accurate because it is only based on the existing structure of the processes. Second, our approach is applicable to processes described in different formats, like pictures and videos, as it does not rely on format-specific techniques like NLP. The main disadvantage of our approach is that it is only applicable to semi-structured resources. We argue that this is not a severe limitation because a large amount of know-how on the Web has some degree of structure. This structure is spontaneously created by Web users as it leads to less ambiguous instructions which are more human understandable. Like the DBpedia project \cite{Auer2007DBpedia}, our system exploits the existing structure of a particular kind of Web repositories to create a large and generic nucleus of Linked Data.

\subsection{Linked Data Integration} \label{subsec:comp2integration}
The representation of human know-how as Linked Data is an important step in making such processes machine understandable. The benefits of this representation, however,  are limited by the amount of knowledge contained in a single set of instructions. Instructions on ``how to apply for a job'', for example, might mention the step ``submit a resume'' without explaining what a ``resume'' is, and how it can be produced. This is a limitation for human users, which might need to search for additional knowledge in order to understand the instructions. This limitation is even more critical for machines, which cannot compensate for the missing knowledge with common sense.
\par Linked Data integration can overcome this limitation by allowing artificial agents to complement the information contained in a single set of instructions with existing related knowledge. Linked Data is the ideal infrastructure for this type of integration, as it allows the creation of links between distributed knowledge sources on the Web. Our framework allows the discovery of two of the most common kinds of links for human know-how. The first kind follows the concepts of inputs and outputs by linking the objects involved in a process with the corresponding DBpedia entities of the same type. The second kind links the steps of a process with other related processes. It should be noted that our Linked Data Integration component creates new links which did not exist before. Unlike our Knowledge Extraction and Representation component, it is not only based on user-generated structure but it utilizes also other techniques, like NLP and Machine Learning, which result in a margin of error. 
\subsubsection{DBpedia Links.} \label{sec:DBpediaLinks} \label{subsec:comp2dbpediaLinks}
Integrating human know-how with DBpedia involves finding links between procedural and declarative knowledge. Two of the most common relations at the intersection of these types of knowledge are the concepts of inputs and outputs. The process ``make a pancake'', for example, can be seen as a process which outputs an object of type ``pancake'' and requires, among others, the ingredient ``milk'' as an input. Our system attempts to identify the DBpedia type of inputs and outputs by analyzing their textual label. The label of each entity is processed by the DBpedia Lookup service\footnote{\url{http://wiki.dbpedia.org/lookup/}} to identify related DBpedia entities. Among those entities, our system chooses the one which has the highest textual similarity with the original label.
\par Input entities are selected among the requirements of a process. Output entities, instead, are selected among the labels of the top-level processes which contain a \emph{creation verb}. A creation verb is verb which semantically implies the creation of its object, such as the verbs ``create'', ``produce'' and ``build''. The object of the creation verb is considered as a candidate output. For example, for the top-level process ``make a pancake'', the word ``pancake'' would be considered as a possible output because the verb ``make'' is a creation verb.
\par The discovery of both input and output types allows the discovery of input/output (I/O) links between processes. An I/O link indirectly connects a process that outputs an entity of a given type with another process that requires an entity of the same type. For example, the same DBpedia entity might be linked to (1) the input ``cover letter'' of the process on ``how to apply for a job'' and (2) the output of the process ``how to write a cover letter''. The combination of these two links forms an I/O link between the two processes.

\subsubsection{Decomposition Links.} \label{subsec:comp2decompositionLinks} 
Step decomposition links are used to connect abstract processes with their steps, steps with their sub-steps and so forth. Within a finite set of instructions, the steps at the bottom of this decomposition hierarchy can be considered \emph{primitive} processes. A primitive process is a process that cannot be decomposed further into sub-processes. Instructions on ``how to apply for a job'', for example, might mention the step ``prepare a resume'' without explaining how this can be achieved. Processes which are not primitive are called \emph{complex}.
\par The execution of a primitive process assumes that the agent following the instructions is able to perform it without requiring any further information. This assumption might be incorrect both for human and artificial agents, thus creating the need to retrieve additional information. In order to exploit related knowledge, our system generates decomposition links between primitive processes and related complex processes. For example, the primitive step ``prepare a resume'' could be linked to a set of detailed instructions on ``how to prepare a resume''.
\begin{figure}[tb]
	\caption{Diagram of the generation of the decomposition links between processes}
			\centering
			\includegraphics[width=1.00\textwidth]{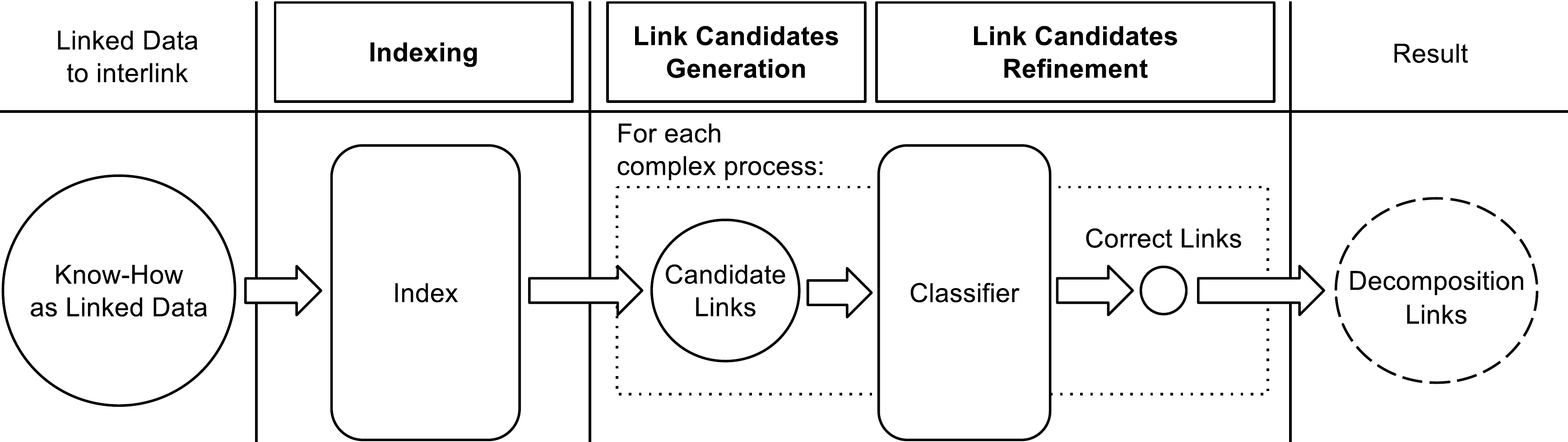}
	\label{fig:DecompositionLinksGeneration}
\end{figure}
\par Our system for generating decomposition links is divided into the three main phases schematized in Figure \ref{fig:DecompositionLinksGeneration}. The first phase involves indexing the textual descriptions of all the processes using an efficient text search engine. This phase addresses the scalability issue caused by the large number of links to consider. 
\par During the second phase, the index is queried to retrieve a small subset of candidate primitive entities to link. These  are chosen on the basis of their textual similarity with the the complex process considered. When analysing the complex process with label ``how to prepare a resume'', for example, we might search the index for primitive entities containing the words ``prepare'' and ``resume''.
\par The third and last phase is meant to refine the set of candidate entities by removing those which are not related with the complex process considered despite having a high text similarity. To do this, a number of features are extracted from each candidate entity with the respect to the complex process considered. These features are then processed by a classifier to decide whether an entity should be linked or not. Examples of the features that can be computed between two entities are the Inverse Document Frequency of the words in common, the number of shared categories, and the number of words in common between the contexts. The context of an entity can be obtained by considering the description of the other entities belonging to the same set of instructions. The keywords ``apply'' and ``job'', for example, could be included in the context of the step ``prepare a resume'' if this step is related to the task ``how to apply for a job''.

\section{Implementation} \label{sec:implementation}

\begin{table}[tp]
	\centering
		\begin{tabular}{ | l | l | l | l |}
			\hline
			& wikiHow & Snapguide & Total \\ \hline
			Number of main processes & 167,232 & 44,464 & 211,696 \\ \hline 
			Total number of entities & 1,871,468 & 737,768 & 2,609,236 \\ \hline
		\end{tabular}
	\caption{Statistics of the knowledge extraction (May 23rd 2014)}
	\label{tab:statistics}
\end{table}

\par To evaluate our framework we have applied it in a large-scale real-world scenario. Our knowledge extraction system analyzed the web pages of the wikiHow and Snapguide websites, two of the largest sources of semi-structured human know-how on the Web. Each web page containing instructions was analyzed and its structure was converted into Linked Data. In total, over 200,000 processes were extracted. More details about this extraction are listed in in Table \ref{tab:statistics}.

\par After extracting the Linked Data representation of a large number of processes, we applied our Linked Data integration system. Our system followed the method described in section \ref{sec:DBpediaLinks} to discover the links between DBpedia entities and the inputs and the outputs of the processes. 
The results of this experiment can be found in Table \ref{tab:statisticsDBpediaIntegration}. 
The precision was manually evaluated separately for the inputs ($P_{I}$) and the outputs ($P_{O}$) on 300 randomly selected links for each type. A link was considered wrong (1) if it linked an entity which was not an input or an output of the process or (2) if the type of the input or output did not correspond to the linked DBpedia type.

\begin{table}[tp]
	\centering
		\begin{tabular}{ | l | l | l | l |}
			\hline
			& Inputs & Outputs & Total \\ \hline
			Number of linked entities & 255,101 & 4,467 & 259,568 \\ \hline 
			Number of different DBpedia types linked & 8,453 & 3,439 & 10,166 \\ \hline 
			Precision & 96\% ($P_{I}$) & 98.3\% ($P_{O}$) & 96\% ($P_{I+O}$) \\ \hline
		\end{tabular}
	\caption{Results of the DBpedia integration experiment}
	\label{tab:statisticsDBpediaIntegration}
\end{table}

\par Lastly, our Linked Data integration system generated decomposition links between different processes. Following the method described in section \ref{subsec:comp2decompositionLinks}, all the entities extracted from human know-how were indexed using the text search engine Apache Lucene.\footnote{\url{http://lucene.apache.org/core/}} 
For each complex and primitive process pair, 34 features were computed and classified by the WEKA \cite{Hall2009WEKA} implementation of a Random Forest classifier. The process pairs classified as correct were linked and added to the set of the discovered decomposition links. The classifier was trained using a manually classified set of 1000 randomly-generated links. 
\par The results of our knowledge extraction and integration experiment are available online through the HowLinks\footnote{\url{http://w3id.org/prohow/main/}} Web application. This application demonstrates how Linked Data can be used to have an integrated visualization of both procedural and declarative knowledge retrieved from different sources.

\section{Evaluation} \label{sec:evaluation}
To understand the significance of the results of our integration experiment, we need to compare them with a relevant benchmark. Since our system is the first to integrate human know-how, we cannot compare our results with a previous experiment. We can however compare them with an existing manual integration effort that is being performed by the wikiHow community. The members of this community, in fact, are not only active in the creation and the refinement of individual sets of instructions, but are also actively creating links between different processes. This community effort can be seen as evidence of the human benefits of know-how integration.
\par To evaluate the results of this community effort we have extracted all the user-generated links found in the description of each wikiHow process. These links have been split into two groups. The first group consists of the links found in the requirements of a process. These links typically connect a required object to a process that can produce such object. As such, they have the same functional role of the I/O links found by our system. The second group is made of the links found in the steps of the process. These links connect a step with another process which provides additional information about that step. As such, they have the same functional role of the decomposition links found by our system. 
\par Having identified a comparable set of links, we proceeded to compute three quality metrics on both results set. These quality metrics are the following:
\begin{itemize}
	\item Precision of the links. Correct links should provide relevant information on how to perform or obtain the linked entity. For example, our system correctly linked the step ``Avoid smoking cigarettes, cigars or pipes around the baby'' with the relevant process ``How to Avoid Smoking''. The wikiHow community integration, instead, linked this step with the irrelevant processes ``How to Smoke a Cigar'' and ``How to Prevent Frozen Water Pipes''.
	\item Number of links found.
	\item Coverage of the links. This metric evaluates the number of processes which contain at least one link to another process. A better integration is achieved when the links are evenly spread between the various entities. 
\end{itemize}
\begin{table}[tp]
	\centering
		\begin{tabular}{ | l | l | l | l | }
			\hline
			                                  & \textbf{WH-C} & \textbf{WH} & \textbf{WH+S} \\ \hline \hline 
				
			Precision of I/O links            & 65\%                      & 94.3\%                    & 94.1\%     \\ \hline 	
			Number of I/O links					      & 4,560                     & 93,883                    & 183,094     \\ \hline 	
			Coverage of I/O links					    & 3,342 (1.9\%)             & 35,169 (21\%)             & 58,029 (27.4\%)    \\ \hline \hline 	
			
			Precision of decomposition links  & 71\%                      & 82.1\%                    & 82.4\%     \\ \hline 	
			Number of decomposition links			& 101,496                   & 127,468                   & 193,701     \\ \hline 	
			Coverage of decomposition links		& 45,250 (27.1\%)           & 69,859 (41.8\%)           & 90,217 (42.6\%)     \\ \hline 	\hline 
			
			Total precision of the links      & 70.7\%                    & 87.3\%                    & 88.1\%     \\ \hline 	
			Total number of links			        & 106,056                   & 221,351                   & 376,795     \\ \hline 	
			Total coverage of the links		    & 45,999 (27.5\%)           & 84,350 (50.4\%)           & 114,166 (53.9\%)    \\ \hline
		\end{tabular}
	\caption{Comparison of the wikiHow community integration (\textbf{WH-C}) with the results of our integration of wikiHow (\textbf{WH}) and of both wikiHow and Snapguide (\textbf{WH+S})}
	\label{tab:finalcomparison}
\end{table}
The result of this comparison can be seen in Table \ref{tab:finalcomparison}. For each of the two types of links generated by the wikiHow community (\textbf{WH-C}), the precision has been evaluated manually on 200 randomly selected links. The precision of the I/O links generated by our system (\textbf{WH+S}) is defined as the probability that both the input and the output links involved are correct: $P_{I/O}=P_{I}*P_{O}$. As such, it can be derived from the precision values shown in Table \ref{tab:statisticsDBpediaIntegration}. The precision of the decomposition links generated by our system (\textbf{WH+S}) is determined by the precision of the classifier used to select them. This precision was evaluated using 10-fold cross validation.
\par It should be noted that the links generated by the wikiHow community only interlink wikiHow resources. On the contrary, our system integrates procedural knowledge both from the wikiHow and the Snapguide repositories. To make a fair comparison, Table \ref{tab:finalcomparison} also shows the evaluation of the links generated by our system which only connect wikiHow resources (\textbf{WH}).
\par The result of our evaluation shows how our automatic approach to human know-how integration significantly outperforms manual community-based integration efforts. This is shown for all the metrics considered and for both types of links. This result demonstrates how our framework can be used to significantly increase the value of human know-how by automatic means. We take this result as strong evidence for the effectiveness of our integration framework.

\section{Conclusion}
This paper has described the first framework for the integration of human know-how into the Linked Data Cloud.
Human know-how is an important source of knowledge on the Web, but its potential applications are limited by a general lack of structure and isolation from other knowledge. 
We have surveyed attempts to overcome these limitations both by user communities, trying to manually add structure and links to human know-how, and by existing research projects, trying to exploit this knowledge to develop intelligent systems. None of these approaches, however, automated the integration of this type of knowledge.
\par We proposed a Linked Data framework to automate both the extraction of human know-how, and its integration with related knowledge. We chose Linked Data as the ideal format to represent distributed knowledge on the Web. 
To validate this framework, we have applied it in a real-world scenario. First, this framework generated the Linked Data representation of over 200,000 processes by extracting human know-how from the wikiHow and Snapguide websites. Lastly, we have used our framework to link these processes both with each other and with DBpedia entities.
We have evaluated the quality of these links and showed how they significantly outperform existing community-based integration efforts. This result demonstrates how the integration of human know-how as Linked Data can immediately benefit Web users by allowing them to access the vast amount of know-how on the Web more efficiently. The application of this knowledge to develop intelligent systems has not been investigated and remains a promising direction for future work.

\bibliographystyle{abbrv}
\bibliography{phdbib} 

\end{document}